\documentclass[letterpaper]{article} 
\usepackage{kkkkkk23} 
\usepackage{times}  %
\usepackage{helvet}  %
\usepackage{courier}  %
\usepackage[hyphens]{url}  %
\usepackage{graphicx} %
\usepackage{natbib}  %
\usepackage{caption} %
\usepackage{algorithm}
\usepackage{algorithmic}
\usepackage{amsmath}
\usepackage{newfloat}
\usepackage{listings}

\DeclareCaptionStyle{ruled}{labelfont=normalfont,labelsep=colon,strut=off} 
\floatstyle{ruled}
\newfloat{listing}{tb}{lst}{}
\floatname{listing}{Listing}

\title{AB2CD: AI for Building Climate Damage Classification and Detection}
\author {
    Maximilian Nitsche\equalcontrib,\textsuperscript{\rm 1,\rm 2}
    S. Karthik Mukkavilli\equalcontrib,\textsuperscript{\rm 3}
    Niklas K\"uhl,\textsuperscript{\rm 4, \rm 1}
    Thomas Brunschwiler\textsuperscript{\rm 3}
}
\affiliations {
    \textsuperscript{\rm 1} IBM Consulting, Germany\\
    \textsuperscript{\rm 2} Karlsruhe Institute of Technology, Germany\\
    \textsuperscript{\rm 3} IBM Research -- Europe, Switzerland\\
    \textsuperscript{\rm 4} University of Bayreuth, Germany\\
    maximilian.nitsche@ibm.com, karthik.mukkavilli@ibm.com, kuehl@uni-bayreuth.de, TBR@zurich.ibm.com
}

\usepackage{bibentry}

\begin{document}

\maketitle

\begin{abstract}
We explore the implementation of deep learning techniques for precise building damage assessment in the context of natural hazards, utilizing remote sensing data. The xBD dataset, comprising diverse disaster events from across the globe, serves as the primary focus, facilitating the evaluation of deep learning models. We tackle the challenges of generalization to novel disasters and regions while accounting for the influence of low-quality and noisy labels inherent in natural hazard data. Furthermore, our investigation quantitatively establishes that the minimum satellite imagery resolution essential for effective building damage detection is 3 meters and below 1 meter for classification using symmetric and asymmetric resolution perturbation analyses. To achieve robust and accurate evaluations of building damage detection and classification, we evaluated different deep learning models with residual, squeeze and excitation, and dual path network backbones, as well as ensemble techniques. Overall, the U-Net Siamese network ensemble with F-1 score of 0.812 performed the best against the xView2 challenge benchmark. Additionally, we evaluate a Universal model trained on all hazards against a flood expert model and investigate generalization gaps across events, and out of distribution from field data in the Ahr Valley. Our research findings showcase the potential and limitations of advanced AI solutions in enhancing the impact assessment of climate change-induced extreme weather events, such as floods and hurricanes. These insights have implications for disaster impact assessment in the face of escalating climate challenges.
\end{abstract}

\section{Introduction}

One of the leading causes of fatalities during a natural hazard is the collapsing of buildings~\citep{rashidian_detecting_2021}. Thus, the rapid assessment of damage to residential buildings and other facilities during or right after a disaster is crucial.

Until today, building damage has been assessed via the manual inspection of aerial images or extensive field surveys~\citep{rashidian_detecting_2021, dong_comprehensive_2013}. Recent advancements in remote sensing have made it easy to acquire enormous volumes of data about nearly every region of the Earth's surface within a few days. Computer vision and deep learning advances further accelerate the trend towards a real-time AI-based assessment of natural hazard-induced building damage.

Since the publication of the xBD dataset~\citep{gupta_creating_2019}, many deep learning architectures tackling the challenge of building damage detection were released~\citep{da_building_2022, weber_building_2020, wu_building_2021, tilon_post-disaster_2020}. This made xBD the leading benchmark for all subsequent satellite building damage detection techniques in research. While literature does evaluate the models’ generalization to hold-out images during training, their analyses do not mimic a real-world application, which requires the hold-out of entire events or they do not provide convincing quantitative statistical analyses about the ability of the model to generalize to unseen events. In this paper, we will improve previous work and address its shortcomings. 

Mono-temporal architectures have been evaluated in \cite{abdi_building_2021} for generalization, training ResNet34 classifiers on the Haiti~earthquake~(2010) and testing the model’s performance on the Woolsey fire (2018). However, since they fine-tuned the model on 20\% of samples provided for the Woolsey event, the model has already adapted to the test event. The performance on two unknown disasters: the Beirut explosion (2020) and Hurricane Laura (2020) were analysed in \cite{wu_building_2021} to verify the transferability of their proposed model. Like xBD, the data originates from the Maxar Open Data Program and has to be labeled manually by the authors. As the satellite imagery stems from the same source, we consider the evaluation of the model’s generalization as an in-distribution test, which does not benchmark the transferability to other remote sensing sources. Furthermore, they only visually inspect the model’s segmentation output for one image per individual disaster. Thus, their analysis lacked a quantitative evaluation based on a statistically significant amount of images. The current benchmarks in the literature show great potential for deep learning architectures to rapidly assess building damage~\citep{abdi_building_2021, weber_building_2020, tilon_post-disaster_2020}.

The main contributions of our work are three-fold: i) we systematically assess the minimum resolution of remote sensing imagery required for computer vision models to work in the real world; ii) we quantitatively evaluate the in-distribution generalization on a statistically significant number of images, lacking in prior work, by performing cross-validation on three subsets of the xBD data set where we leave out several events for robustness of testing in the real world; iii) we finally test our models’ transferability to an out-of-distribution event, namely the Ahr Valley flood 2021 in Germany, with a statistically significant number of images being evaluated. 

We tested our models on public datasets from German Aerospace Center for Ahr Valley 2018 \url{https://www.geoportal.rlp.de} and xBD from \url{https://xview2.org/dataset}.

\section{Methodology}

Our model architecture is split into two stages: localization and classification of building damage. We first train a U-Net~\citep{ronneberger_u-net_2015} to output a building localization mask based on the pre-event image only. This initial training scheme sensitizes the model to detect buildings without wasting costly damage labels. In the second and last stage of our training scheme, the localization U-Net of the first stage is adopted to a Siamese Network~\citep{koch_siamese_2015} with shared weights between the pre- and post-event input. The outputs resulting from the two forward passes through the U-Net are concatenated and fused by a convolutional layer. The final output of the Siamese network is of size $1024*1024*5$ (width $*$ height $*$ channels), where the first channel of the last axis corresponds to the general building detection and the remaining channels to the four ordinal damage levels of the xBD data set: no damage, minor, major damage, and destroyed. The final damage class is determined via the weighted average along the channel axis being above a given threshold.

To improve the generalization and general performance of the proposed architecture, we do heavy model ensembling with backbones of different sizes and architecture principles, e.g. residual learning (He et al., 2016). Overall, we implement four different backbones which all share the same aforementioned architecture. The four pre-trained backbones comprise a ResNet \citep{he_deep_2016} (ResNet34), a Squeeze-and-Excitation Network (SENet154), a ResNeXt~\citep{xie_aggregated_2017} that employs Squeeze-and-Excitation blocks~\citep{hu_squeeze-and-excitation_2018} (SE-ResNeXt50), and a Dual Path Network~\citep{chen_dual_2017} (DPN92). We train for each backbone, three models with differently initialized seeds. All backbones’ weights are pre-trained on the large-scale ImageNet data set~\citep{deng_imagenet_2009}.

Besides model ensembling, we augment the image pairs to improve model robustness with regards to different day time satellite passings of the affected regions, different sources of remote sensing data, and view angles of the satellites. The image augmentation techniques comprise classical image flipping, rotation, shifting, cropping, and change of hue. Moreover, we add Gaussian noise, blur and change saturation, brightness, as well as contrast. As an optimizer, we choose the adaptive gradient algorithm Adam  \citep{kingma_adam_2017} with $L_2$ regularization and weight decay regularization (AdamW) as proposed by \citeauthor{loshchilov_decoupled_2019}.
The loss between model output $p_i$ and target map $g_i$ for pixel $i$ is defined as the weighted combination of the Dice $L_{\text{Dice}}$ \citep{cardoso_generalised_2017} and Focal loss $L_{\text{Focal}}$ \citep{lin_focal_2020} with $\gamma=2$. The focal loss adopts the standard cross entropy criterion such that for $\gamma>0$ it puts an emphasis on relatively hard-to-classify examples by reducing the loss of properly classified examples.

\begin{align}
L_{\text{Dice}} &= \frac{2 \sum_i^N p_i g_i}{\sum_i^N p_i^2+\sum_i^N g_i^2} \\
L_{\text{Focal}}(r_i) &= -\left(1-r_{i}\right)^\gamma \log \left(r_{i}\right) \\
r_{i} &= (1-g_i) \cdot (1-p_i) + g_i \cdot p_i
\end{align}

The xBD data set suffers a great class imbalance since the three damage classes minor, major damage, and destroyed only account for one-fourth of the overall buildings labeled. Thus, we oversample all image pairs with damaged buildings. Additionally, we oversample the classes 'minor' and 'major damage' since these are particularly difficult to distinguish. The models are initially trained and evaluated following the split of the xBD data set. 
The model architecture is implemented in Python and the deep learning library PyTorch \citep{paszke_pytorch_2019}. We train all models on two GPUs of type NVIDIA Tesla V100 SXM2 32 GB.

\section{Results}

First, we evaluate the performance of our model and the various enhancement techniques on the xBD test set. Additionally, we investigate model behavior and performance for a few of the individual disaster types. The model’s behavior is further explored in the subsequent analysis of resolution where we determine the minimal spatial resolution required for localizing and classifying building damage at different levels of detail. Finally, we present the outcomes of our generalization study which is split into two parts: the analysis of the in-distribution (re-splitting of the xBD data set) and out-of-distribution (real-world application for the Ahr Valley event) generalization gap of the proposed model.

\begin{figure}[h]
  \centering \includegraphics[width=\linewidth]{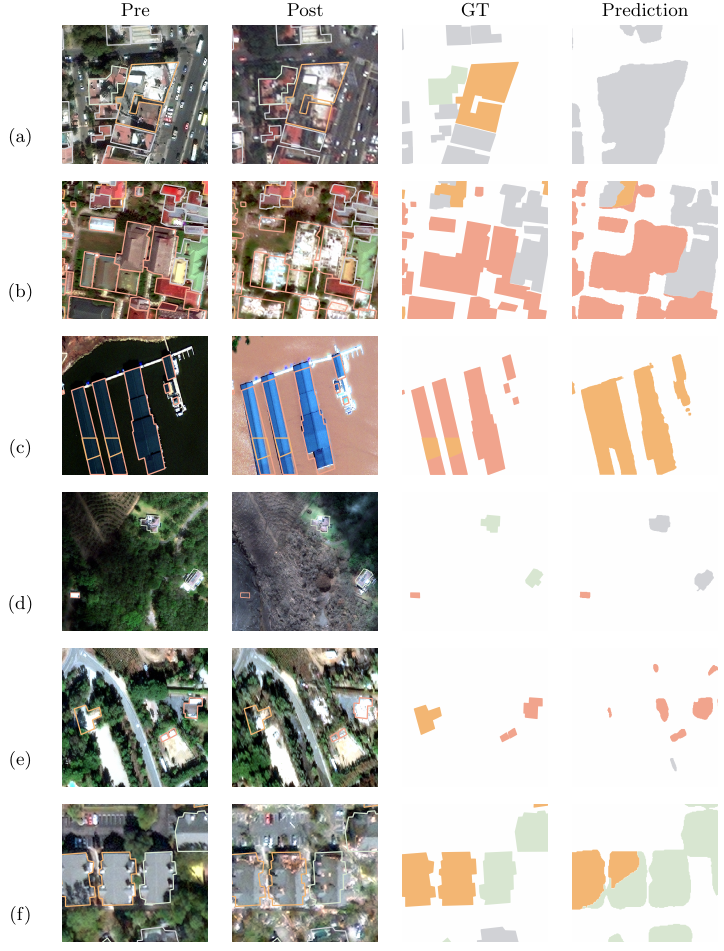}
  \caption[Examples for the ensemble model segmentation results of xBD samples from the test set.]{Examples for the ensemble model segmentation results alongside the pre- and post-imagery as well as ground truth (GT) of xBD samples from the test set: (a) Mexico earthquake (2017), (b) Palu tsunami (2018), (c) Midwest flooding (2019), (d) Guatemala volcano (2018), (e) Socal fire (2017), (f) Hurricane Michael (2018). The color of the building footprints and prediction output corresponds to the assigned damage level: unclassified (blue), no damage (grey), minor damage (green), major damage (orange), and destroyed (red).}
  \label{fig:figure1}
\end{figure}

Table~\ref{tab:table1} shows the performance of our model and the different backbones along the evaluation metrics defined in the Methodology. The ensemble of models achieves an overall challenge score of 0.8119 with a localization F1-score of 0.8624 and classification macro-average F1-score of 0.7897. According to the classification F1-scores for the individual damage levels, minor ($F1^{E}_{C_2}=0.6444$) and major ($F1^{E}_{C_3} =0.7859$) damaged buildings are significantly more difficult to classify and harder to separate from the other classes. In contrast, the classification of extreme cases, where buildings are left unaffected ($F1^{E}_{C_1}=0.9234$) or entirely destroyed ($F1^{E}_{C_4}=0.8640$), are easily detected and well segmented by the model. 
The single ResNet34 U-Net has the same relations of performance measures but performs generally worse than the ensemble. But the heavy ensembling of the four model architectures especially causes a great improvement for the two difficult-to-classify classes: minor and major damage, as the F1-scores increased by 0.0520 and 0.0208, respectively. The building footprints are equally well segmented by the single ResNet34 model ($\Delta{F1_{loc}}=F1^{E}_{loc} - F1^{Res34}_{loc}=0.0037$). Examples of the segmentation results of the ensemble of models are depicted in Figure~\ref{fig:figure1}.

\begin{table*}
\centering

\caption[Performance overview of model architectures with different backbones.]{Performance overview of model architectures with different backbones. The best-performing F1-scores of our models implemented are underlined. The xBD benchmark is the best-performing model on the leader-board of the xView2 challenge~\citep{defense_innovation_unit_xview2_2019} which is the first place solution we used.}\label{tab:table1}
    
\begin{tabular}{llcccccccc}
\hline
&& \multicolumn{1}{l}{\textbf{Localization}} & \multicolumn{6}{c}{\textbf{Damage Classification}} & \multicolumn{1}{l}{\textbf{Score}} \\
\hline
&&& No damage & Minor & Major & Destroyed & Binary & Macro-avg. &  \\
\hline
\textbf{xBD benchmark} & F1 & 0.8635 & 0.9234 & 0.6444 & 0.7859 & 0.8640 & - & 0.7898 & 0.8119\\
\hline
Siamese U-Nets & F1 & \underline{0.8624} & 0.9234 & \underline{0.6444}	& \underline{0.7859} & 0.8640 & \underline{0.8816} & \underline{0.7897} & \underline{0.8119} \\
Ensemble & Precision & 0.7983 & 0.9682 & 0.5925 & 0.7811 & 0.9304 & 0.8642 &&  \\
& Recall & 0.9377 & 0.8918 & 0.6330 & 0.7805 & 0.8273 & 0.8997 &&  \\
\hline
ResNet34 & F1        &        0.8587 &     0.9212 &        0.5924 &        0.7651 &     0.8657 &  0.8712 &      0.7640 & 0.7924 \\
& Precision &        0.7981 &     0.9641 &        0.5860 &        0.7618 &     0.9286 &  0.8566 &             &        \\
& Recall    &        0.9293 &     0.8821 &        0.5990 &        0.7684 &     0.8108 &  0.8862 &             &        \\
\hline
SENet154 &F1        &        0.8595 &     0.9256 &        0.5673 &        0.7664 &     0.8702 &  0.8707 &      0.7551 & 0.7865 \\
& Precision &        0.7930 &     0.9610 &        0.5481 &        0.7967 &     0.9136 &  0.8651 &             &        \\
& Recall    &        0.9383 &     0.8928 &        0.5878 &        0.7383 &     0.8307 &  0.8763 &             &        \\
\hline
SE-ResNeXt50 &F1        &        0.8579 &     0.9606 &        0.6096 &        0.7784 &     \underline{0.8882} &  0.8723 &      0.7856 & 0.8072 \\
& Precision &        0.8346 &     0.9641 &        0.5757 &        0.7847 &     0.9174 &  0.8622 &             &        \\
& Recall    &        0.8876 &     0.9571 &        0.6479 &        0.7722 &     0.8608 &  0.8827 &             &        \\
\hline
DPN92 & F1        &        0.8538 &     \underline{0.9575} &        0.6116 &        0.7814 &     0.8873 &  0.8664 &      0.7865 & 0.8066 \\
& Precision &        0.8475 &     0.9679 &        0.5622 &        0.7614 &     0.9323 &  0.8381 &             &        \\
& Recall    &        0.8832 &     0.9473 &        0.6705 &        0.8026 &     0.8465 &  0.8966 &             &        \\
\hline
\end{tabular}
\end{table*}

To further understand the errors of our model, we evaluated confusion matrices for the xBD test set and the individual natural hazard types, which demonstrated that the model can distinguish between damage grades while class confusion primarily occurs between adjacent classes. Moreover, we had lower F1-scores for minor and major damage as the most confusion originated from classifying minor as no damage (23.53\%) and major as minor damage (20.76\%). The model is not robustly calibrated to correctly classify minor flood damage and additionally confuses destroyed buildings with unaffected ones. We conclude that flood damage grades are harder to classify and distinguish. This finding is supported by the visual inspection for hurricanes, for instance, which heavily impact the roof of a building, whereas water-related natural hazards flood the construction and a major part of the damage is caused inside, leaving the roof’s appearance unaffected. Therefore, flood damage seems to be harder to detect in nadir remote sensing data.

Based on the previous finding that building damage can vary depending on the natural hazard type, we tested whether disaster type-specific models outperform the model trained on all kinds of events. We refer to the models which are designated for one of the six disaster types as expert models. We denote the model trained on all events as the universal model. Table~\ref{tab:table2} shows the performance of both models being tested on all flooding events in the test set. This also applies to hurricanes that resulted in flooding, such as Hurricane Michael (2018).

\begin{table*}
\centering
\caption[Performance of the universal vs flood expert ResNet34 for the flooding events in the test set of xBD.]{Performance of the universal vs flood expert ResNet34 for the flooding events in the test set of xBD. The universal model was trained on all types of natural hazards. The flood expert was fine-tuned on flooding events only. The performance measures are the localization $F1_{loc}$ and macro-average classification F1-score $F1_{cls}$. The best-performing variation is underlined. Classification F1-score $F1_{C_l}$ with $l\in\{1,2,3,4,b\}$, where 1: No damage, 2: Minor damage, 3: Major damage, 4: Destroyed, b: Binary damage}\label{tab:table2}
\begin{tabular*}{\textwidth}{@{\extracolsep{\fill}}lcccccccc@{}}
\hline
\textbf{Model} & $F1_{loc}$ & $C_1$ & $C_2$ & $C_3$ & $C_4$ & $C_b$ & $F1_{cls}$\\
\hline
Universal & \underline{0.8530} & \underline{0.8344} & 0.5415 & 0.7308 & 0.4619 & \underline{0.8475} & 0.6080\\
Flood expert & 0.8489 & 0.7985 & \underline{0.5634} & \underline{0.7906} & \underline{0.5022} & 0.8279 & \underline{0.6366}\\
\hline
\end{tabular*}
\end{table*}

Overall, the results indicate differences in performance between the two models. First, buildings are localized with higher accuracy by the universal model as their general appearance does not deviate between disaster types. The same finding holds true for the classification of buildings that are not damaged. However, the results indicate that differentiating between the three levels of damage is better captured by the hazard expert model as it outperforms the universal model for all three levels of damage ($C_l, l\in\{2,3,4\}$). This also leads to the improved overall classification of flood-induced building damage. 

\subsection{Image Resolution Analysis}\label{sec:img_res_analysis}

There are various sources of remote sensing data that are free of charge and have a high revisit time. But these data sources often suffer a low spatial resolution while very-high resolution data is still expensive and not publicly available~\citep{kresse_change_2022}. Thus, we aim to find the minimum image resolution at which building damage can still be confidently assessed.

To determine the minimum required satellite image resolution, we gradually downsample the input images and run inference of the model on the entire test set of xBD. 
For this investigation, we only derive the performance considering the ResNet34 model, due to its similar performance compared to the ensemble of models (see Table \ref{tab:table1}). 

\subsubsection{Symmetric resolution perturbation}
\begin{figure}
    \centering \includegraphics[width=0.5\textwidth]{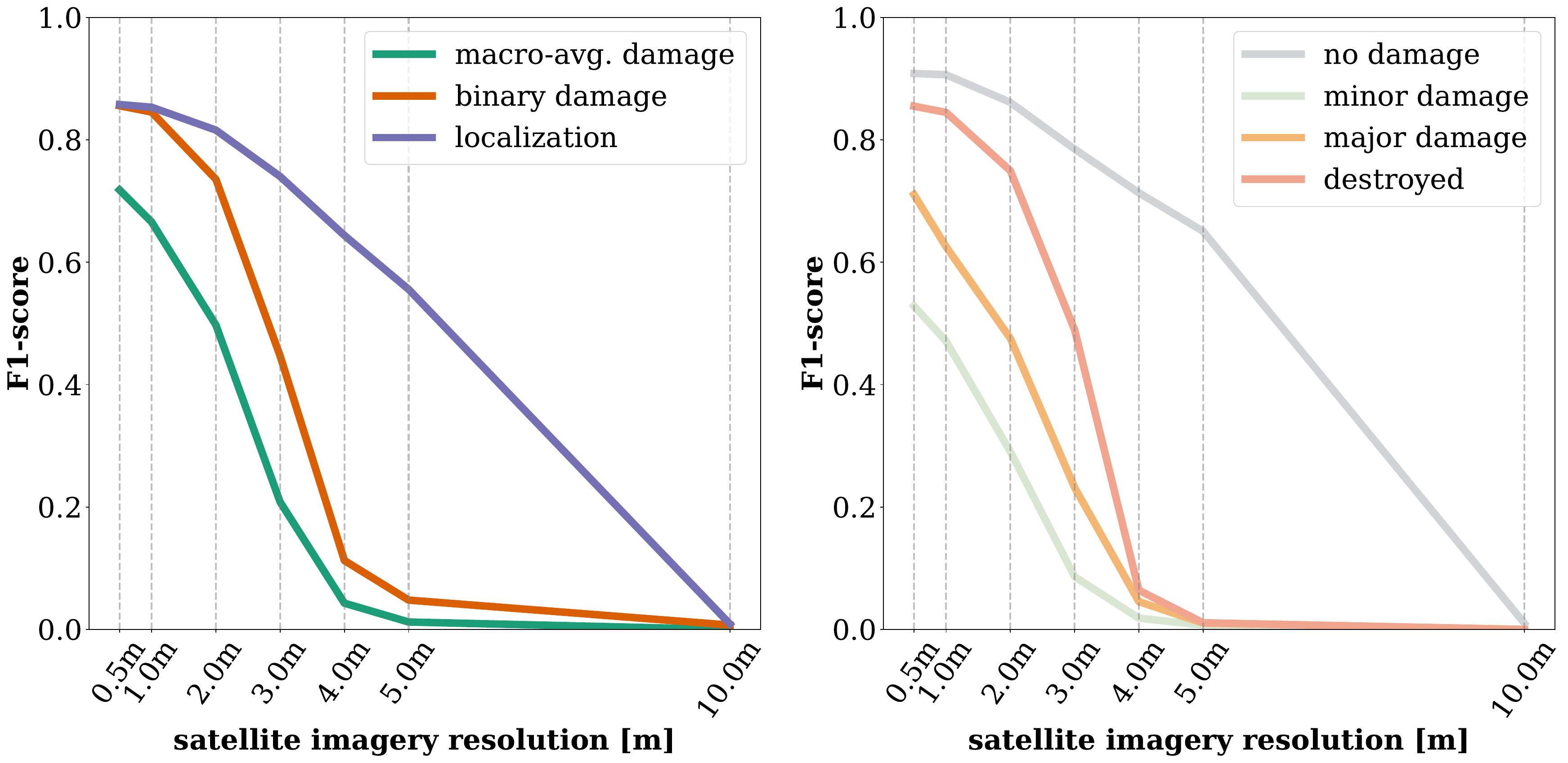}
    \caption{Symmetric resolution analysis: Building damage estimation for the simultaneous perturbation of satellite pre- and post-images.}
    \label{fig:figure2}
\end{figure}

We first decrease resolution $r$ of both pre- and post-event images simultaneously and track the performance of building localization $F1_{loc}\left(r\right)$, binary damage $F1_{C_b}\left(r\right)$, and macro-average damage F1-scores $F1_{cls}\left(r\right)$ in Figure \ref{fig:figure2}. We plot the decrease along the schedule of 0.5 (original image resolution), 1.0, 2.0, 3.0, 4.0, 5.0, and 10.0 meters resolution. 

Notably, the model's ability to localize building footprints prevails until an image resolution of 10 meters---and decreases much slower than the damage classification curves. However, localization performance is only acceptable until a resolution of 5 meters with an F1-score of 0.55.

In contrast to the localization performance, the damage classification curves are much steeper. Here every meter of resolution improvement matters. Both the binary and macro-average damage classification F1-scores are almost zero for resolutions below 4 meters. The performance drop especially takes place for the resolution perturbation within 2.0 and 4.0 meters. When we compare the binary damage F1-score with the one averaged over all four damage classes, we observe a similar curve progression of the classification performance drop. But the binary damage F1-score is constantly higher than the macro-average damage classification. As a result, we can follow that for the general detection of building damage an image resolution of at least 2 meters is required ($F1_{C_b}(2m) = 0.73$). 

For the rating of damage at higher detail as in the joint damage scale of xBD, however, information at 1-meter resolution or higher is needed to still confidently distinguish between the four damage levels ($F1_{C_1}(1m) = 0.90$ and $F1_{C_4}(1m) = 0.84$). Since minor and major damage was already difficult to differentiate in the test set of original quality ($F1_{C_2}(0.5m) = 0.52$, $F1_{C_3}(0.5m) = 0.71$), a higher resolution should be preferred for these damage levels. 

\subsubsection{Asymmetric resolution perturbation} 
\begin{figure}
    \centering \includegraphics[width=0.5\textwidth]{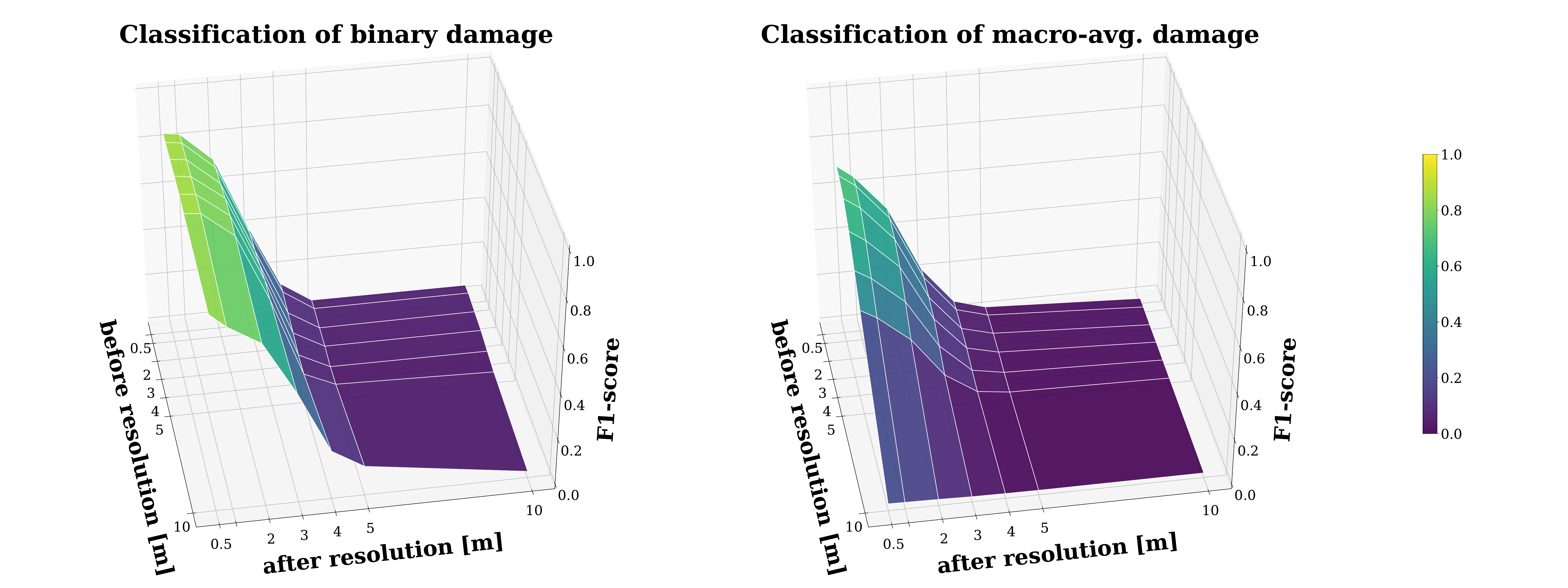}
    \caption{Building damage for different asymmetric resolutions: perturb resolution of the pre \& post satellite imagery individually}
    \label{fig:figure3}
\end{figure}

Besides the general resolution requirements for satellite data, we are interested in the interplay of the pre- and post-event images in order to explore the underlying model behavior.
Similarly to our previous symmetric resolution analysis, we gradually scale down both of the original xBD image pairs of 0.5-meter resolution up to a maximum of 10 meters. However, we perturb the images individually by tracking performance decrease above every combination of resolutions within the range of $r \in \{0.5, 1, 2, 3, 4, 5, 10\}$ meters. Consequently, we end up with 49 combinations of input image resolutions to test the model's performance. The performance frontiers of binary $F1_{C_b}(r_{pre}, r_{post})$, and macro-average damage classification $F1_{cls}(r_{pre}, r_{post})$, as well as building localization $F1_{loc}(r_{pre}, r_{post})$ are shown in Figure \ref{fig:figure3}. Note that the diagonal of the asymmetric resolution performance frontier corresponds to the performance of our previous symmetric resolution perturbation, as pre- and post-event resolution is the same.
Interestingly, the binary damage detection does not heavily depend on the pre-event imagery as the performance frontier in Figure \ref{fig:figure3} is rather constant along the pre-event image resolution axis. However, Figure \ref{fig:figure3} shows that the comparison between pre- and post-event imagery is necessary to assess building damage at a more detailed scale, e.g., the joint damage scale defined in the xBD data set. While performance depends primarily on the post-event resolution, it also drops along the decreasing pre-event resolution. We further investigated the interplay of the co-registered images for individual damage levels. The performance surface of the three damage classes ’minor damage’, ’major damage’, and ’destroyed’ follow our previous findings that minor and major damage is harder to classify than destroyed buildings and the classification relies on the comparison of both input images at high resolution. However, the classification performance of not damaged buildings prevails even for a post-event resolution of 10 meters.

The performance frontier shown in Figure \ref{fig:figure3} of building localization reflects our training scheme since performance only drops when the pre-event imagery is decreased in resolution. This indicates the model’s robustness since the localization of buildings is also based on damage classification which in turn heavily depends on the post-event image, as outlined above.

 \subsection{Generalization Analysis} 
 Finally, we tested the model’s generalization with regard to events never seen during training. We split the analysis into two parts: the in-distribution and out-of-distribution testing. The in-distribution testing is evaluating the generalization for unknown events but from the same data set and, therefore, the same distribution. The out-of-distribution analysis is, besides the unknown event, also testing for the generalization to other remote sensing sources.

 \subsubsection{Event Cross-Validation} 
For the in-distribution testing of generalization, we perform cross-validation on the event level in xBD i.e., we leave out events from the training and afterwards evaluate the model’s performance on these events. But as the xBD data set comprises 19 natural disasters, performing full-blown leave-one-out cross-validation would be too computationally expensive. Thus, we follow the data split of~\cite{benson_assessing_2020} and create the following three folds to test generalization: 
\begin{enumerate}
    \item Fold: Pinery bushfire, Joplin tornado, and Sunda tsunami
    \item Fold: Moore tornado and Portugal wildfire
    \item Fold: Lower Puna volcano, Tuscaloosa tornado, and Woolsey fire
\end{enumerate}

We train and validate the ResNet34 model on all disaster types but the ones in the corresponding fold. Afterward, we evaluate the newly trained model on the unseen events in the fold and report the performance measures in Table~\ref{tab:table3}.

\subsubsection{Robust evaluation in the wild (Ahr Valley Flood case)}

After we observed only a minor generalization gap for the damage localization and classification for the in-distribution setting, we test for generalization out-of-distribution. Thus, we initially run an inference of the model trained on disasters in xBD for half of the Ahr Valley data. While the xBD data set is recorded from the Maxar remote sensing satellite at 0.5 meters resolution, the Ahr Valley data stems from high-resolution airborne remote imagery, and was downsampled to the same coarser resolution.

\begin{figure}[h]
    \centering \includegraphics[width=\linewidth]{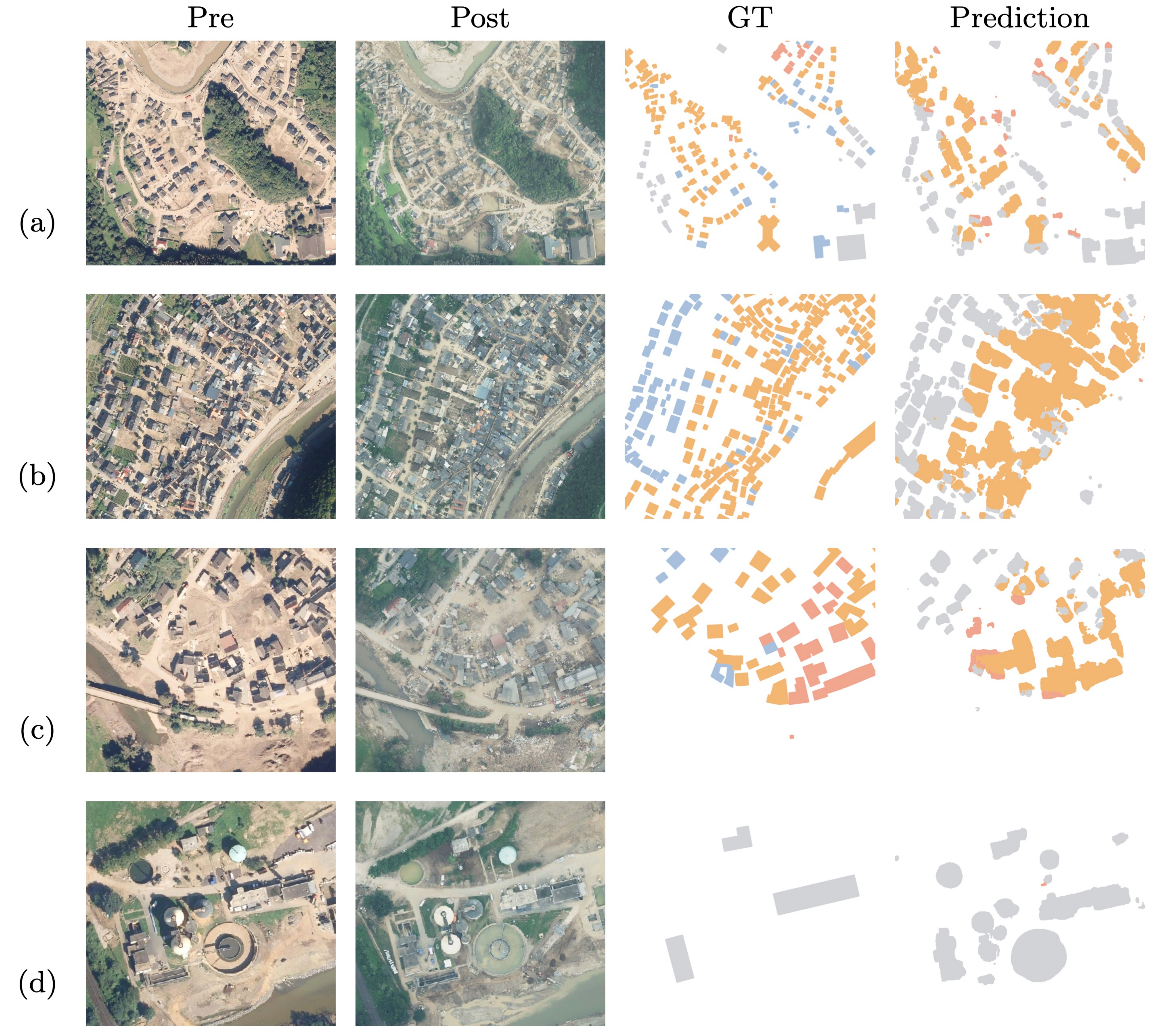}
    \caption{Examples of segmentation results alongside the pre- and post-event imagery (Landesamt f\"ur Vermessung Geobasisinformationen Rheinland-Pfalz (LVG), 2021) as well as the ground truth (GT) of the Ahr Valley: (a) exemplary segmentation for all damage levels available, (b) good segmentation, (c) destroyed buildings misclassified as ’damaged’, and (d) silos misclassified as buildings. The color of the building footprint corresponds to the assigned damage level: unclassified (blue), possibly damaged (grey), damaged (orange), or destroyed (red).}
    \label{fig:figure4}
\end{figure}

As the Ahr Valley data has no visual overlap between the image pairs, we can randomly split the data into training and test sets without inducing data leakage. According to the evaluation metrics shown in Table~\ref{tab:table4}, the model is able to localize and classify undamaged buildings in the Ahr Valley ($F1_loc=0.5462$ and $F1_{C_1}=0.7321$). However, when it comes to the classification of damaged or even destroyed buildings, the model's performance drops significantly to almost zero ($F1_{C_{2/3}}=0.0142$ and $F1_{C_4}=0.0308$). Consequently, the overall macro-avg. classification of building damage as the harmonic mean across F1-scores is almost zero as well ($F1_{cls}=0.0107$). We focus on the binary classification of damage in the Ahr Valley. In view of this, the model has an F1-score of $F1_{C_{b}}=0.2521$. Besides the quantitative evaluation, Figure~\ref{fig:figure4} depicts a series of exemplary predictions of the ResNet34 for the Ahr Valley event.

In the second step, we aim for tracking the adaptation of the model to the new data over the number of training samples provided. Therefore, we define the adaptation $A(s)$ of our model being trained on a share $s\in[0,0.5]$ of the Ahr Valley data as the following: 

\begin{align}
A_{loc}(s) &= F1_{loc}(s) - F1_{loc}(s=0) \label{eq:loc_adop}\\
A_{cls}(s) &= F1_{cls}(s) - F1_{cls}(s=0) \label{eq:cls_adop}\\
A_{C_l}(s) &= F1_{C_l}(s) - F1_{C_l}(s=0) \label{eq:level_cls_adop}
\end{align}

We start by fine-tuning the model on the maximum of available training data i.e., half of the Ahr Valley data ($s=0.5$) which was not used for testing in the previous step. The performance of the model fine-tuned on the Ahr Valley event is depicted in the last row (ResNet34 s=0.5) of Table~\ref{tab:table4}. 
The results indicate that the model only adapts to a limited extent. While the classification of buildings labeled as 'damaged' adapts considerably with $A_{C_{2/3}}(0.5)=0.1980$, the overall ability to differentiate between damaged and intact buildings remains unchanged as the adaptation is $A_{C_{b}}(0.5)=0.0138$. 

\begin{table*}
\centering
\caption[Comparison of the ResNet34 segmentation performance between the original xBD test set and the three folds for cross-validation.]{Comparison of the ResNet34 segmentation performance between the original xBD test set and the three folds for cross-validation. We test the ResNet34 model on the three folds after training it on the remaining events. The last row is the average performance across the folds. \footnotemark[1]{Classification F1-score $F1_{C_l}$ with $l\in\{1,2,3,4,b\}$, where 1: No damage, 2: Minor damage, 3: Major damage, 4: Destroyed, b: Binary damage}
\footnotemark[2]{Pinery bushfire, Joplin tornado, and Sunda tsunami.}
\footnotemark[3]{Moore tornado and Portugal wildfire.}
\footnotemark[4]{Lower Puna volcano, Tuscaloosa tornado, and Woolsey fire.}}\label{tab:table3}
\setlength{\tabcolsep}{1pt} 
\begin{tabular*}{\textwidth}{@{\extracolsep{\fill}}lcccccccc@{}}
\hline
\textbf{Model} &$F1_{loc}$ &\footnotemark[1]$C_1$ & $C_2$ & $C_3$ & $C_4$ & $C_b$ & $F1_{cls}$\\
\hline
xBD test & 0.8635 & 0.9234 & 0.6444	& 0.7859 & 0.8640 & 0.8816 & 0.7897\\
1st~fold\footnotemark[2] & 0.8778 &     0.9264 &        0.6733 &        0.5970 &     0.8600 &  0.8985 &     0.7404\\
2nd~fold\footnotemark[3] & 0.8984 &     0.9479 &        0.5409 &        0.4959 &     0.8107 &  0.8234 &     0.6500\\
3rd~fold\footnotemark[4] & 0.8711 &     0.9191 &        0.6840 &        0.4336 &     0.8395 &  0.8621 &     0.6614\\
Fold avg. & 0.8824 & 0.9311 & 0.6327 & 0.5088 & 0.8367 & 0.8613 & 0.6839\\
\hline
\end{tabular*}
\end{table*}

\begin{table*}
\centering
\caption[Comparison between the pre-trained and fine-tuned ResNet34 segmentation performance on the Ahr Valley flooding 2021 in Germany.]{Comparison of the ResNet34 segmentation performance on the Ahr Valley flooding 2021 in Germany. The first block of rows is the performance of the model which was trained on the xBD training set only and was not fine-tuned on a share $s\in[0,0.5]$ of the Ahr Valley data ($s=0$). The last block shows the evaluation of the ResNet34 which was trained on the xBD training set and additionally fine-tuned on half of the Ahr Valley data ($s=0.5$). \footnotemark[1]{Classification $C_l$ for damage level with $l\in\{1,2,3,4,b\}$, where 1: Possibly damaged, 2/3: Damaged, 4: Destroyed, b: Binary damage}}\label{tab:table4}
\begin{tabular*}{\textwidth}{@{\extracolsep{\fill}}llccccccc@{\extracolsep{\fill}}}
\hline
&&\multicolumn{1}{l}{\textbf{Localization}} & \multicolumn{4}{c}{\textbf{Damage classification}\footnotemark[1]} &\\
\hline
\textbf{Model} &\textbf{Metric}&&$C_1$ & $C_{2/3}$ & $C_4$ & $C_b$ &$F1_{cls}$&  \\
\hline
ResNet34 & F1 & 0.5462 & 0.7321 & 0.0142 & 0.0308 & 0.2521 & 0.0107\\
$s=0$ & Precision & 0.4355 & 0.8093 & 0.2150 &  0.0307 &  0.3854 &\\
& Recall & 0.7325 & 0.6683 & 0.0074 & 0.0308 & 0.1873\\
\hline
ResNet34 & F1 & 0.5378 & 0.7348 & 0.2122 & 0.0036 & 0.2659 & 0.0288\\
$s=0.5$ &Precision & 0.4274 & 0.8173 & 0.5068 & 0.0030 & 0.3965\\
&Recall & 0.7252 & 0.6675 & 0.1342 & 0.0048 & 0.2000\\
\hline
\end{tabular*}
\end{table*}

\section{Discussion}

In our research, we analyzed the performance and generalization of our model in various settings. We initially find that the model performs well for the classification of damage on different levels of detail for the xBD test set. But the performance metrics and confusion matrices indicated that the model is challenged to differentiate between partially damaged buildings and to separate them from the extreme cases where buildings are destroyed or left unaffected.

Furthermore, we started to investigate whether natural hazard-induced building damage varies for the different types of disasters. Owing to the previous finding that 'minor' and 'major damage' is difficult to differentiate, we primarily built our investigation on the performance of our model for these damage levels. We find that the visual appearance of damage might vary as our model confuses different pairs of damage levels for the two hazard types: fire and flood. It's particularly intriguing that the model classifies buildings destroyed by a flood as unaffected. However, the model's confusion captures the fact that floods often do not damage the roof of a building but rather the interior. This finding challenges the model used in this work as convolutional operations only extract local information, while attention-based architectures consider the global context, such as the surrounding water of a flood~\citep{hao_attention-based_2020}. In addition, data sources beyond nadir images might be more important for a comprehensive understanding of the scene~\citep{bommasani_opportunities_2022}. 

To tie in with the aforementioned findings, we uncover a demand for expert models specifically trained for a given type of natural hazard in our studies. While a universal (i.e. hazard type agnostic) model is sufficient for the detection of damage of any severity, expert models might lead towards an accurate damage assessment at a more granular level. However, this finding is for the time being limited to our ablation studies on floods.

In the first question, we ask for the minimum resolution of remote sensing imagery required to confidently assess building damage. We answer this question in a bifurcated manner since the minimum resolution depends on the level of detail damage being assessed. If the objective is to classify a building as either damaged or not damaged, our model requires an image resolution higher than 3 meters per pixel. Interestingly, we find in our asymmetric resolution analysis that this spatial resolution criterion only applies to the post-event imagery. Our analysis indicates that the detection of damage in a binary decision setup is independent of the pre-event imagery. This finding supports current research efforts which concentrate on mono-temporal techniques whenever building damage is not further split into different levels of damage~\citep{abdi_building_2021}. This finding is very relevant, as pre-event images are typically available from satellites only. However, for major natural disasters, detailed airborne post-event analysis can be expected.

On the other hand, when it comes to the classification of damage on a scale of different grades, our model requires an even higher image resolution of 1 meter or below. This requirement for the remote sensing data is in line with our previous finding that the detection of partially-damaged buildings is the most challenging for our model. While the resolution limit primarily applies to the post-event imagery, our asymmetric analysis implies a strong dependence between the resolutions of the two input images. Thus, for the estimation of damage into different levels the comparison of pre- and post-event imagery appears to be requisite. This finding complements the increasing rise of multi-temporal architectures in literature and other research~\citep{dong_comprehensive_2013}.

Moreover, the damage might not only be specific to the type of disaster itself, as we have pointed out above, but also varies on the individual event level. Since the xView2 challenge splits the imagery of a disaster across the different subsets for training and testing, any model could have already adapted to the visual appearance of damage for a specific event. 

In our second question, we aim for evaluating the ability of models to generalize to unknown events and regions. Thus, we split the data along disasters and test for in-distribution generalization of our model. We find the generalization gap of our model for detecting building damage to be smaller than what is found for similar architectures in literature~\citep{benson_assessing_2020}. The data suggest that for the binary classification of building damage, there is no generalization gap as performance only drops by 2\%. Whereas binary detection prevails across the three folds, the detection of major damage drops significantly. This result argues for the visual representation of partial damage to be even specific to an individual disaster. But further analysis is required to reach a final decision on this matter.

In addition, we analyzed the generalization to an out-of-distribution setting where the region of the disaster, as well as the source of the remote sensing data, is different. The ResNet34 model pre-trained on xBD signals a large generalization gap for the Ahr Valley event 2021 in Germany. But the model does not even adapt to the disaster when trained on half of the data. We attribute the poor results to the quality of the labels provided by Copernicus Emergency Management Service (CEMS). This finding stems from the fact that there is no class that explicitly labels buildings as unaffected by the flood. Moreover, following a visual inspection of the data buildings are either not annotated at all or labeled in batches. In terms of the out-of-distribution generalization, we can, therefore, not conclude any result. However, the noisy CEMS labels indicate the current limitations of impact assessments and the need for model advances as presented here.

Overall the model shows the ability to generalize for natural hazard events and regions never seen during training. However, the model's generalization to unknown events and its robustness against the image resolution is primarily limited by the level of detail at which building damage has to be assessed. 

\section{Conclusion and Outlook}\label{sec:conclusion_and_outlook}

In this work, we assess the robustness of current deep learning for the detection and classification of natural hazard-induced building damage in remote sensing data. For this purpose, we implement the first-place solution of the well-known xView2 challenge. To test robustness, we perturb the input images and track the model's performance over different spatial resolutions. We further contribute to both theory and practice by evaluating the ability of the model to generalize for natural hazard events never seen during its training phase. 

As a result of the resolution analysis, we find that our implemented model is able to differentiate between damaged and unaffected buildings for post-event images higher than 3 meters in resolution and does not consider the pre-event imagery. However, for the classification of damage into different grades, the comparison of pre- and post-event images at resolutions below 1 meter per pixel seems to be needed. Therefore, the decision towards a mono- or multi-temporal approach should depend on the level of detail building damage needs to be extracted. The extracted range of spatial resolution for our deep learning model might guide future research to focus on remote sensing sources which meet these requirements.

The generalization analysis initially builds on the repartitioning of the xBD data set at the disaster level and evaluates the model on events not used during training. We observe only a negligibly small generalization gap for the three test folds. Moreover, we create our own data set for the Ahr Valley Flood 2021 in Germany and apply our model to this event. While the first insights indicate a large generalization gap for this disaster, we cannot finally conclude any result due to the low quality and granularity of the labels provided by current post-disaster assessments.
Furthermore, our studies indicate that the visual expression of different building damage levels is at least hazard type specific but may also be unique for the individual disasters as suspected by~\cite{tilon_post-disaster_2020}. This preliminary comparison between a hazard-type agnostic model and an expert model specifically trained for floods supports the recent rise of large foundation models which are designated to be fine-tuned for specific downstream tasks and domains~\citep{bommasani_opportunities_2022}. 

Finally, the remote sensing-based classification of building damage into different grades aims toward the rapid and accurate estimation of direct and  tangible losses. Thus, future research should focus on the translation of the physical damage extracted by deep learning models into financial projections. A promising field of research lies ahead.

\section{Acknowledgements}
 We would like to thank the DaVis Team of the Federal Office of Civil Protection and Disaster Assistance and the German Aerospace Center for generously sharing the aerial images taken of the Ahr Valley flood 2021 in Germany.

\bibliography{kkkkkk23}

\end{document}